\documentclass[submission,copyright,creativecommons]{eptcs}
\providecommand{\event}{AREA 2020} 

\usepackage[T1]{fontenc}
\usepackage[utf8]{inputenc}
\usepackage[english]{babel}
\usepackage[inline, shortlabels]{enumitem}
\usepackage{comment,csquotes,graphicx,hyperref,lscape,multirow,subcaption,url,xcolor,gensymb}

\begin{document}

\title{Testing the Robustness of AutoML Systems}
\author{Tuomas Halvari \and Jukka K. Nurminen \and Tommi Mikkonen
\institute{Department of Computer Science, University of Helsinki, Finland \email{first.[initial.]last}@helsinki.fi}}

\def\authorrunning{T. Halvari et al.}
\def\titlerunning{Testing the Robustness of AutoML Systems}

\maketitle

\begin{abstract}

Automated machine learning (AutoML) systems aim at finding the best machine learning (ML) pipeline that automatically matches the task and data at hand. We investigate the robustness of machine learning pipelines generated with three AutoML systems, TPOT, H2O, and AutoKeras. In particular, we study the influence of dirty data on accuracy, and consider how using dirty training data may help create more robust solutions. Furthermore, we also analyze how the structure of the generated pipelines differs in different cases.  


\end{abstract}

\section{Introduction}\label{sec:intro}

Automated machine learning (AutoML) systems are used to find the best machine learning (ML) pipeline matching the task and data at hand, typically classification or regression. This includes model selection and hyperparameter optimization. Finding good models and hyperparameters are hard and time-consuming tasks for human experts, and they frequently involve a lot of trial-and-error experimentation. The promise of AutoML is that computers can automate these repetitive tasks and come up with good pipelines with little human effort. The drawback is that AutoML systems require a lot of computing power and the quality of the results varies. A recent overview of different AutoML systems \cite{truong2019towards} echoes these issues. 

In this paper, we investigate the robustness of ML pipelines produced by AutoML mechanisms. Our focus is on user-friendly AutoML systems, which do not require prior knowledge about the data, algorithm choices, or hyperparameter spaces. For critical use cases, it is not enough that AutoML produces pipelines with accurate inference results. It is also important that the resulting pipelines tolerate faults, e.g. Gaussian noise, in data. At the moment AutoML is in an early phase and there seem to be no prior studies focusing on the robustness of their results. As AutoML gains maturity and ML systems are applied in safety-critical tasks deeper understanding of the robustness of the resulting systems is important. This paper is an early step towards that direction. 

When building AI systems for robots and other autonomous devices, one consideration is their robustness against unexpected inputs, which commonly occur as a result of hardware or other problems in sensing and communication. Another class of unexpected inputs, which is outside of the scope of the present paper, is adversarial attacks, which aim for minimum input changes able to confuse the ML algorithms. 

Recently there have been many papers comparing the performance of different AutoML systems \cite{balaji2018benchmarking,truong2019towards,gijsbers2019open}. Likewise, several papers discuss robust training of neural networks and vulnerability to adversarial inputs \cite{aleks2017deep,Shaham_2018,moosavidezfooli2015deepfool,Weng2019}. Our study combines these two perspectives.  More precisely, we measure the robustness of three different AutoML systems (TPOT \cite{OlsonGECCO2016}, H2O AutoML \cite{H2OAutoML}, and AutoKeras \cite{jin2019auto}) with artificial inputs where we can control the type and amount of faults in the training and testing data. Our focus is on how dirty data, which arise if e.g. the camera of a robot is tilted or the lens is covered with dust, affects the performance. In particular, we focus on the following questions:

\begin{itemize}

\item How accurate are the AutoML generated models when testing with dirty data (Section \ref{sec:results_clean_accuracies}).

\item What effect would training with on-purpose dirty data have on model accuracy? (Section \ref{sec:results_dirty_accuracies}).

\item How similar/different pipelines do the different AutoML tools produce in the above cases? (Sections \ref{sec:results_clean_pipelines} and \ref{sec:results_dirty_pipelines})

\item How do the results vary as a function of the amount of faults in the testing data? (Sections \ref{sec:results_clean} and  \ref{sec:results_dirty})

\item What AutoML systems to recommended for different use cases? (Section \ref{sec:results_recommendations})

\end{itemize}


In the scope of this paper, we study the three AutoML systems (TPOT, H2O, AutoKeras) in the presence of data faults in training and/or testing data. Our experiments are built using dpEmu fault injector framework \cite{Nurminen2019}, which makes running such experiments easy. We control the amount of faults in the data and use two of the fault sources provided by dpEmu, namely Gaussian noise and rotation. Our testing focuses on image classification tasks. We use six different data fault levels with each data fault source and two different image datasets, namely Digits \cite{scikit-learn} and Fashion \cite{xiao2017fashion}.


The structure of the paper is as follows. Section \ref{sec:automl} gives background of AutoML systems. It also introduces the AutoML systems we study in this paper and the criteria for their selection. In Section \ref{sec:methodology}, we discuss how the measurements were conducted and describe the used datasets and the faults generated to them. The results are presented in Section \ref{sec:results} and their meanings discussed in Section \ref{sec:disc}. Finally, we present our conclusions in Section \ref{sec:conclusions}.

\section{AutoML systems}\label{sec:automl}

AutoML systems are meta-level machine learning algorithms, which use other ML solutions as building blocks for finding the optimal ML pipeline. In this context, an ML pipeline means the set of algorithms and their hyperparameters that the ML system uses to infer results from data. An AutoML system has to consider multiple ML pipelines and search values for their parameters. It needs to optimize each candidate pipeline to an adequate level but also ensure that enough time and resources are used to experiment with alternative pipelines. As a result, using AutoML systems can consume a lot of computing resources. 


Typical tasks that many AutoML systems support are classification and regression. In various examples and benchmarks, typically image or text data are used. Some AutoML system like AutoKeras \cite{jin2019auto} even offer specialized image and text classifiers. Image data is usually easy to handle as a pixel array, with an integer value for each pixel, is used to represent each image. Pretty much all classification systems support this kind of input out of the box, and not much preprocessing is required. Unfortunately, this is not the case with text data, as it comes in many shapes. One dataset might be a list of strings and another a preprocessed dataset, where each string is represented as a sequence of integers representing the overall frequency in the data. While some AutoML systems like TPOT \cite{OlsonGECCO2016} and H2O AutoML \cite{H2OAutoML} accept numerical arrays as inputs and do not care what the numbers represent, for example, AutoKeras has only specialized classifiers for image and text data. Because AutoKeras's text classifier only accepts text data as a list of strings and uses a built-in preprocessor, fair comparison to other more general AutoML systems may prove difficult. Therefore, we have left out the text data and only focus on image recognition in our study. 

In this paper we study three different AutoML systems: TPOT, H2O, and AutoKeras.  These were chosen because:

\begin{itemize}

\item

All three provide a simple Python API and basic use requires only a few lines of code, which makes them easy to include in our benchmarks.

\item

These three are different enough in their approach to constructing the optimal ML pipeline. They also use different ML library backends.

\item

Unlike some of the available AutoML systems, with these three no previous knowledge of the data is required and no search space for the models and hyperparameters needs to be specified making them easy tools also for casual users.

\end{itemize}

Other free to use and open-source AutoML systems include MLBox \footnote{\url{https://github.com/AxeldeRomblay/MLBox}} with required user-defined search spaces, auto-sklearn \cite{Feurer2019}, which is similar to TPOT, and TransmogrifAI \footnote{\url{https://transmogrif.ai/}}, which uses Apache Spark. Widely used cloud providers, such as Google Cloud, support AutoML \footnote{\url{https://cloud.google.com/automl/}}, but, as part of the cloud business model, they are usually closed source and not free to use.

\subsection{TPOT}\label{sec:automl_tpot}

TPOT \cite{OlsonGECCO2016} is a tree-based pipeline optimization tool. It uses the scikit-learn library \cite{scikit-learn} as the ML backend, and the classification models used include several models (Naive Bayes, Random Forest, Gradient Boosting, Linear SVC, Logistic Regression, etc.) from scikit-learn and the XGBoost classifier \cite{Chen_2016}. Aside from the actual ML models, the pipelines that TPOT creates can contain for example scalers, feature selection techniques, dimensionality reduction techniques, and other preprocessors \cite{OlsonGECCO2016}. TPOT uses genetic programming to evolve the pipeline sequence and hyperparameters to optimize certain criteria, like classification accuracy \cite{OlsonGECCO2016}. Inspecting the code reveals that TPOT uses predefined hyperparameter spaces for each model it considers. \footnote{\url{https://github.com/EpistasisLab/tpot/blob/master/tpot/config/classifier.py}} TPOT offers no GPU support.

\subsection{H2O}\label{sec:h2o}

H2O AutoML \cite{H2OAutoML} is a small and new part of the H2O.ai ML platform \footnote{\url{https://www.h2o.ai/}}. H2O's core code is written in Java, but a Python API is also provided. H2O AutoML supports the training of Stacked Ensemble models, which are collections of individual models. These Stacked Ensemble models are constructed by a meta learner called Super Learner \cite{super-learner} with a goal of combining a diverse set of different, base or optimized, models together.

The base models that H2O supports are Generalized Linear Models (GLM), Distributed Random Forests (DRF), XGBoost, Gradient Boosting Machines (GBM), and Deep Learning (NN). The hyperparameters used are chosen from a predefined search space using grid search. It seems that H2O chooses from 3 different options. It may use just one of the base models or their hyperparameter-optimized versions. It can also choose a Best Of Family Stacked Ensemble model, which includes one model from each category. The last option available is the All Models Stacked Ensemble pipeline, which can be very long. These three make up quite different choices for the best pipeline and in case of an easy dataset, with high accuracy.

Unlike the other two AutoML systems, H2O uses its own backend, which runs as a Java process. H2O offers a very limited GPU support: only XGBoost models can be trained with GPU, others are limited to CPU. 

\subsection{AutoKeras}\label{sec:autokeras}

AutoKeras builds a deep learning neural model for your task and data. It optimizes both architecture and hyperparameters using neural network morphism guided by Bayesian optimization to select the most promising operations at each stage \cite{jin2019auto}. First, in each stage, the underlying model is trained with the proposed architecture and its performance is measured. Then, a new architecture is generated by optimizing an acquisition function. Finally, the performance of the new architecture is evaluated by training and testing the actual neural network. It uses Tensorflow, Keras and Torch backends. While TPOT and H2O do not support GPUs, AutoKeras offers full GPU support.

One of the interesting features of AutoKeras image classifier is the option to augment the train data to prevent overfitting and possibly increase robustness. \cite{jin2019auto} It uses random crops, random horizontal flips, and cutouts for data augmentation.

\section{Research Approach}\label{sec:methodology}

For performing the measurements, we have used parts of the dpEmu framework \cite{Nurminen2019}, a software framework for emulating common problems in data, testing the robustness of ML systems, and visualizing the results. The essential idea is that the system generates artificial faults to datasets according to predefined or user-defined fault models. The script run.py, that is used to run our benchmarks can be found in our repository \cite{repo}. It first creates different versions of the dataset at different data fault levels, given the data fault source. Then the benchmarked model is trained with different versions of the training data in a loop and after each step, the model is tested with all versions of the test data.  

A total of twelve benchmarks were run for each model ranging from 15 mins to 6 hours. For the Digits dataset, the six benchmarks were 15 min, 30 min and 1 hour for both data fault sources. For the Fashion dataset, the six benchmarks were 1 hour, 3 hours and 6 hours for both data fault sources. This is the time available for each AutoML system to find and train the optimal classification pipeline.
Notice that the images in the Digits dataset are too small in resolution for AutoKeras, so these results are unavailable.

\subsection{Test setup}\label{sec:setup}

All the CPU-only benchmarks used in our testing were run on the University of Helsinki's Kale cluster using Intel Xeon E5-2680 v4 CPU's and a total of 40 cores with more than enough RAM. The benchmarks utilizing GPUs were run with a single Nvidia Tesla v100 GPU.

For TPOT and H2O we used the latest version available at the time of writing. For AutoKeras we used a slightly older but stable version $0.4.0$, which may not have all the features of the newer versions, but enables us to set a time limit to benchmark the system properly with the others. The particular versions for the key components of our test system were: Python $3.7.0$, Java $11.0.2$, AutoKeras $0.4.0$, H2O $3.28.0.3$, TPOT $0.11.1$, XGBoost $1.0.1$, CUDA $10.0.130$ and cuDNN $7.5.0.56$.

\subsection{Datasets}\label{sec:datasets}

Two datasets, Digits and Fashion, of different sizes were used. The smaller dataset is the Digits dataset \cite{scikit-learn}. It consists of 1797 8x8 grayscale images of handwritten digits. The pixel values fall in range $0,\ldots,16$. It was chosen because it is lightweight enough even for the heavier AutoML systems enabling them to optimize the pipelines more instead of just struggling to find a decent solution. 
The larger dataset is the Fashion-MNIST \cite{xiao2017fashion} dataset, consisting of 70000 28x28 grayscale images of Zalando's articles. The pixel values are in the range $0,\ldots,255$. It has been created as a more difficult replacement for the famous MNIST dataset, mainly because MNIST classification is too easy for modern ML algorithms. Like the Digits dataset, it contains images of articles from 10 different classes, 7000 each. With Digits and Fashion, $1/4$ and $1/7$ of the dataset were reserved for testing and the rest for training, respectively.

The key idea is to compare the large Fashion dataset with its big number of good training images with the small set of Digit training data. We especially want to see how fast the smaller training data becomes useless because of the lack of good training images when the amount of faults in the data increases.

\subsection{Data fault sources}\label{sec:data_faults}

\begin{figure}[!ht]
    \centering
    \begin{subfigure}[b]{0.45\textwidth}
        \centering
        \includegraphics[width=\textwidth]{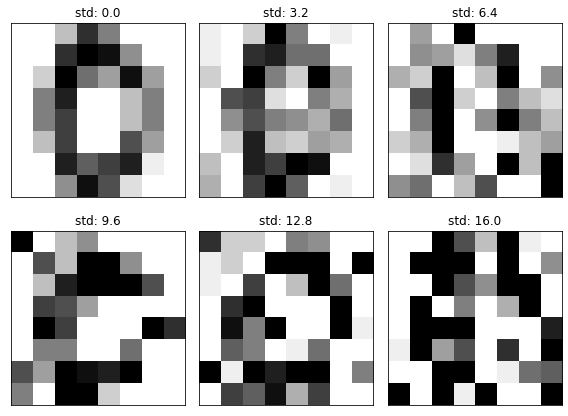}
        \caption{Digits image with different levels of Gaussian noise}
        \label{fig:digits_data_fault_noise}
    \end{subfigure}
    \begin{subfigure}[b]{0.45\textwidth}
        \centering
        \includegraphics[width=\textwidth]{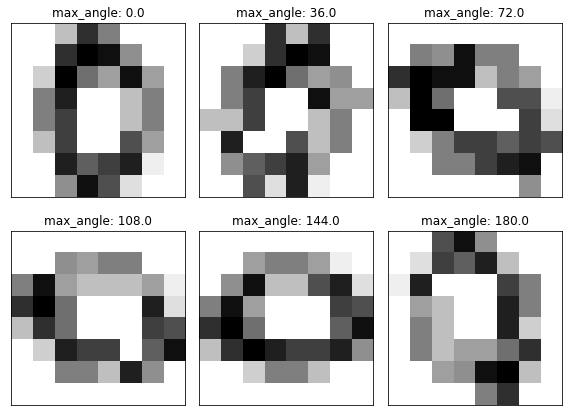}
        \caption{Digits image with different levels of random rotation}
        \label{fig:digits_data_fault_rotation}
    \end{subfigure}
    \hfill
    \begin{subfigure}[b]{0.45\textwidth}
        \centering
        \includegraphics[width=\textwidth]{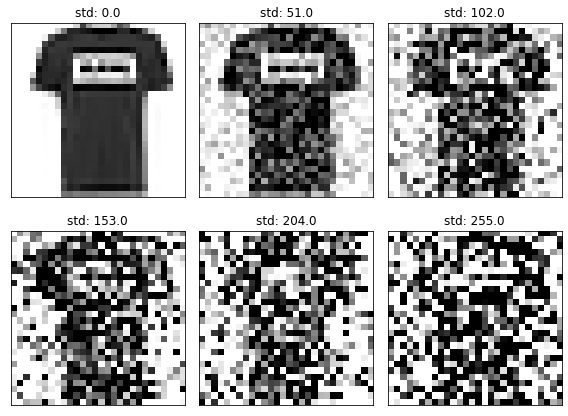}
        \caption{Fashion image with different levels of Gaussian noise}
        \label{fig:fashion_data_fault_noise}
    \end{subfigure}
    \begin{subfigure}[b]{0.45\textwidth}
        \centering
        \includegraphics[width=\textwidth]{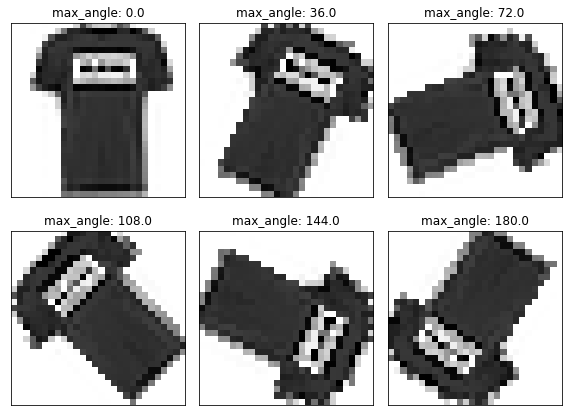}
        \caption{Fashion image with different levels of random rotation}
        \label{fig:fashion_data_fault_rotation}
    \end{subfigure}
    \caption{Example images from Digit and Fashion datasets using different data fault sources at different data fault levels}
    \label{fig:data_fault_examples}
\end{figure}

We used both Gaussian noise and image rotation as data fault types. Figure \ref{fig:data_fault_examples} shows examples of both fault types for both datasets. Both sources generate random faults. This means that even at high error levels it is possible but highly unlikely to get near original images. Notice also that the ranges of pixel values are different in the two datasets as described in Section \ref{sec:datasets}. Six predefined data fault levels are used for both noise and rotation, including the clean level 0, with standard deviation and maximum angle as the data fault parameters respectively.

The reason for choosing these two was that while Gaussian noise effectively destroys parts of the information about the true label from the image, whereas, at least for the human eye, rotating the image makes little difference to the shape of its object.

\subsection{Metrics}\label{sec:metrics}

Our primary metric for image classification is the accuracy score when comparing predicted and true labels. The accuracy score was chosen over the F1-score because there are no imbalanced classes in either of the datasets.

\subsection{Validity and limitations}\label{sec:validity_limitations}

To make the comparison fair, the following points have been considered: 
\begin{itemize}

\item

We focus on image classification because, unlike text processing, it is done in rather similar ways in all three systems. Text data would have required preprocessing for TPOT and H2O while AutoKeras would automate it.

\item

We conduct the testing using time limit based categories in a way similar to other studies \cite{gijsbers2019open}. Otherwise, the runtimes would vary greatly and the results would be more dependent on the default parameters.

\item

We allocated all computing resources to work on one model with one fault type at a time.

\item

We allowed only some of the initialization parameters to be fixed. Such parameters include time limits, random seeds, the parameters enabling the model to use more CPUs or RAM, and the parameters used to modify logging output.

\item

AutoKeras benefits significantly from the GPU use and has a optional feature for image augmentation. Thus we used three different versions for AutoKeras, namely CPU, GPU and GPU with image augmentation.

\end{itemize}

The accuracy of AutoKeras GPU version results varied a lot form one run to the other in comparison to the other tested systems. Therefore, we report the accuracy results for AutoKeras as average over two runs for each benchmark.

Though the two image dataset are quite different regarding to both image and dataset sizes, they are still quite similar with single item in each image. In a future work, a more realistic dataset with colors could also be included. 

\section{Results}\label{sec:results}

In this Section, unless otherwise mentioned, for noise source std is fixed at the second level, meaning $6.4$ for the Digits dataset and $102$ for the Fashion dataset. For rotation source maximum angle is fixed at the last data fault level corresponding to 180 degrees for both datasets, meaning all possible rotations are equally likely. Complete results are available at our GitHub repository \cite{repo}.

\subsection{How good are AutoML generated models with clean training data?}\label{sec:results_clean}

\subsubsection{Accuracies}\label{sec:results_clean_accuracies}

Let us start by comparing the five AutoML systems when both training and testing is done with clean data. The accuracy results for both datasets can be found in Table \ref{tab:accuracies_summary}. They report the maximum accuracy of the six runs with different maximum execution times (typically longer execution times improved the results but not always, see Tables \ref{tab:accuracy_time_clean_summary} and \ref{tab:accuracy_time_dirty_summary}). 

\begin{table}[]
\centering
\caption{Summary of best accuracy results per model for both datasets, given the training and testing data fault sources at a fixed data fault level.}
\label{tab:accuracies_summary}
\resizebox{0.9\textwidth}{!}{%
\begin{tabular}{|l|l|l|l|l|l|l|l|l|l|l|l|l|l|l|}
\hline
Dataset                 & \multicolumn{7}{c|}{Digits}                           & \multicolumn{7}{c|}{Fashion}                          \\ \hline
Training data &
  \multicolumn{3}{c|}{Clean} &
  \multicolumn{2}{c|}{Noise} &
  \multicolumn{2}{c|}{Rotation} &
  \multicolumn{3}{c|}{Clean} &
  \multicolumn{2}{c|}{Noise} &
  \multicolumn{2}{c|}{Rotation} \\ \hline
Testing data &
  Clean &
  Noise &
  Rotation &
  Clean &
  Noise &
  Clean &
  Rotation &
  Clean &
  Noise &
  Rotation &
  Clean &
  Noise &
  Clean &
  Rotation \\ \hline
AutoKeras CPU           & -     & -     & -     & -     & -     & -     & -     & 0.914 & 0.205 & 0.250 & 0.836 & 0.819 & 0.820 & 0.807 \\ \hline
AutoKeras GPU           & -     & -     & -     & -     & -     & -     & -     & 0.925 & 0.283 & 0.244 & 0.833 & 0.812 & 0.839 & 0.829 \\ \hline
AutoKeras GPU with Aug. & -     & -     & -     & -     & -     & -     & -     & 0.945 & 0.159 & 0.278 & 0.744 & 0.851 & 0.881 & 0.877 \\ \hline
H2O                     & 0.987 & 0.676 & 0.289 & 0.973 & 0.842 & 0.887 & 0.838 & 0.905 & 0.449 & 0.233 & 0.838 & 0.821 & 0.805 & 0.792 \\ \hline
TPOT                    & 0.987 & 0.887 & 0.373 & 0.951 & 0.838 & 0.891 & 0.853 & 0.882 & 0.492 & 0.236 & 0.801 & 0.782 & 0.760 & 0.760 \\ \hline
\end{tabular}%
}
\end{table}

Let's for now focus just on the columns with clean training and testing data. For the Digits dataset, it seems that both H2O and TPOT are equally good. On the other hand, with the Fashion dataset, all AutoKeras versions seem strong and data augmentation seems to help with accuracy. H2O seems to beat TPOT slightly. 

\begin{table}[]
\centering
\caption{Summary of the effect of time to accuracy when both training and testing with clean data.}
\label{tab:accuracy_time_clean_summary}
\resizebox{0.7\textwidth}{!}{%
\begin{tabular}{|l|l|l|l|l|l|l|}
\hline
                        & \multicolumn{3}{c|}{Digits} & \multicolumn{3}{c|}{Fashion} \\ \hline
Benchmark & \multicolumn{1}{c|}{15 min} & \multicolumn{1}{c|}{30 min} & \multicolumn{1}{c|}{1 h} & 1.5 h & 3 h & 6 h \\ \hline
AutoKeras CPU           & -       & -       & -       & 0.887    & 0.912   & 0.912   \\ \hline
AutoKeras GPU           & -       & -       & -       & 0.908    & 0.921   & 0.916   \\ \hline
AutoKeras GPU with Aug. & -       & -       & -       & 0.928    & 0.933   & 0.930   \\ \hline
H2O                     & 0.984   & 0.986   & 0.982   & 0.902    & 0.902   & 0.905   \\ \hline
TPOT                    & 0.985   & 0.985   & 0.987   & 0.876    & 0.879   & 0.882   \\ \hline
\end{tabular}%
}
\end{table}

The results for the effect of benchmarking time to the accuracy when both training and testing with clean data can be seen in Table \ref{tab:accuracy_time_clean_summary}. When looking at the accuracy transitions from 1.5 h to 3 h with the larger Fashion dataset, AutoKeras CPU and AutoKeras GPU with image augmentation seem to require more time to reach the optimal performance, when compared to the other three test cases. Especially the CPU version of AutoKeras seems to struggle in creating a neural network with CPU resources only. When testing, AutoKeras CPU with image augmentation enabled did not seem viable at all so GPU training seemed to be the only option.

\begin{figure}[!ht]
    \centering
    \begin{subfigure}[b]{0.45\textwidth}
        \centering
        \includegraphics[width=\textwidth]{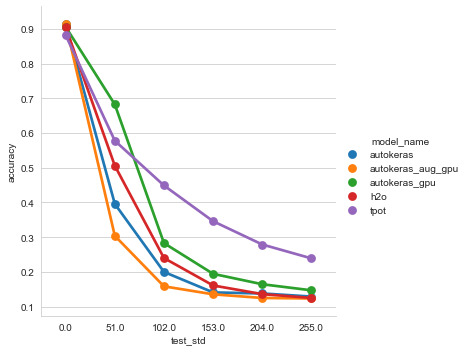}
        \caption{With noise as the data fault source and clean training data.}
        \label{fig:accuracy_plot_noise_clean_dirty}
    \end{subfigure}
    \begin{subfigure}[b]{0.45\textwidth}
        \centering
        \includegraphics[width=\textwidth]{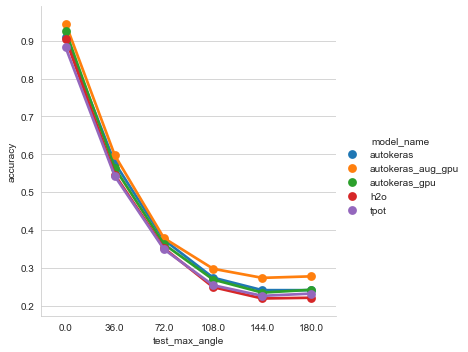}
        \caption{With rotation as the data fault source and clean training data.}
        \label{fig:accuracy_plot_rotation_clean_dirty}
    \end{subfigure}
    \hfill
    \begin{subfigure}[b]{0.45\textwidth}
        \centering
        \includegraphics[width=\textwidth]{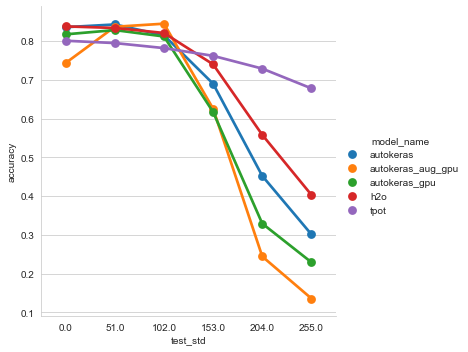}
        \caption{With noise as the data fault source and dirty training data.}
        \label{fig:accuracy_plot_noise_dirty_dirty}
    \end{subfigure}
    \begin{subfigure}[b]{0.45\textwidth}
        \centering
        \includegraphics[width=\textwidth]{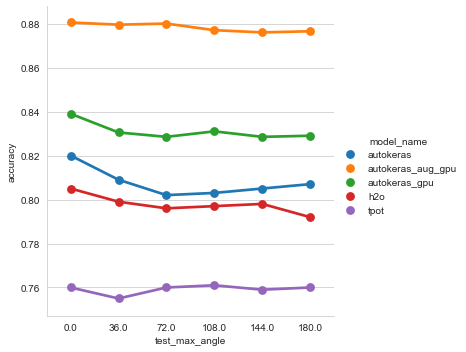}
        \caption{With rotation as the data fault source and dirty training data.}
        \label{fig:accuracy_plot_rotation_dirty_dirty}
    \end{subfigure}
    \caption{Accuracy plots for the 6 h benchmark with the Fashion dataset when testing at different data fault levels given the data fault source and training data.}
\end{figure}

\begin{table}[]
\centering
\caption{Summary of the effect of time to the best pipelines per model given the training data. The results in parentheses are from alternative runs.}
\label{tab:pipelines_summary}
\resizebox{0.9\textwidth}{!}{%
\begin{tabular}{|l|l|l|l|l|l|l|l|}
\hline
 &
  \multirow{2}{*}{\begin{tabular}[c]{@{}l@{}}Training\\ data\end{tabular}} &
  \multicolumn{3}{l|}{Digits} &
  \multicolumn{3}{l|}{Fashion} \\ \cline{1-1} \cline{3-8} 
 &
  &
  15 min &
  30 min &
  1 h &
  1.5 h &
  3 h &
  6 h \\ \hline
\multirow{3}{*}{AutoKeras} &
  Clean &
  - &
  - &
  - &
  \begin{tabular}[c]{@{}l@{}}CNN\\ 16 (16) layers\end{tabular} &
  \begin{tabular}[c]{@{}l@{}}CNN\\ 66 (66) layers\end{tabular} &
  \begin{tabular}[c]{@{}l@{}}CNN\\ 67 (67) layers\end{tabular} \\ \cline{2-8} 
 &
  Noise &
  - &
  - &
  - &
  \begin{tabular}[c]{@{}l@{}}CNN\\ 16 layers\end{tabular} &
  \begin{tabular}[c]{@{}l@{}}CNN\\ 66 layers\end{tabular} &
  \begin{tabular}[c]{@{}l@{}}CNN\\ 67 layers\end{tabular} \\ \cline{2-8} 
 &
  Rotation &
  - &
  - &
  - &
  \begin{tabular}[c]{@{}l@{}}CNN\\ 16 layers\end{tabular} &
  \begin{tabular}[c]{@{}l@{}}CNN\\ 66 layers\end{tabular} &
  \begin{tabular}[c]{@{}l@{}}CNN\\ 483 layers\end{tabular} \\ \hline
\multirow{3}{*}{\begin{tabular}[c]{@{}l@{}}AutoKeras\\ GPU\end{tabular}} &
  Clean &
  - &
  - &
  - &
  \begin{tabular}[c]{@{}l@{}}CNN\\ 69 (483,483,483)\\ layers\end{tabular} &
  \begin{tabular}[c]{@{}l@{}}CNN\\ 68 (69,70,77)\\ layers\end{tabular} &
  \begin{tabular}[c]{@{}l@{}}CNN\\ 68 (70,483,483)\\ layers\end{tabular} \\ \cline{2-8} 
 &
  Noise &
  - &
  - &
  - &
  \begin{tabular}[c]{@{}l@{}}CNN \\ 68 (69) layers\end{tabular} &
  \begin{tabular}[c]{@{}l@{}}CNN\\ 483 (483) layers\end{tabular} &
  \begin{tabular}[c]{@{}l@{}}CNN\\ 80 (483) layers\end{tabular} \\ \cline{2-8} 
 &
  Rotation &
  - &
  - &
  - &
  \begin{tabular}[c]{@{}l@{}}CNN\\ 483 (483) layers\end{tabular} &
  \begin{tabular}[c]{@{}l@{}}CNN\\ 71 (483) layers\end{tabular} &
  \begin{tabular}[c]{@{}l@{}}CNN\\ 77 (79) layers\end{tabular} \\ \hline
\multirow{3}{*}{\begin{tabular}[c]{@{}l@{}}AutoKeras\\ GPU\\ with Aug.\end{tabular}} &
  Clean &
  - &
  - &
  - &
  \begin{tabular}[c]{@{}l@{}}CNN\\ 66 (66,66,66)\\ layers\end{tabular} &
  \begin{tabular}[c]{@{}l@{}}CNN\\ 69 (69,70,483)\\ layers\end{tabular} &
  \begin{tabular}[c]{@{}l@{}}CNN\\ 72 (79,483,483)\\ layers\end{tabular} \\ \cline{2-8} 
 &
  Noise &
  - &
  - &
  - &
  \begin{tabular}[c]{@{}l@{}}CNN\\ 66 (66) layers\end{tabular} &
  \begin{tabular}[c]{@{}l@{}}CNN\\ 66 (67) layers\end{tabular} &
  \begin{tabular}[c]{@{}l@{}}CNN\\ 483 (483) layers\end{tabular} \\ \cline{2-8} 
 &
  Rotation &
  - &
  - &
  - &
  \begin{tabular}[c]{@{}l@{}}CNN\\ 66 (66) layers\end{tabular} &
  \begin{tabular}[c]{@{}l@{}}CNN\\ 483 (483) layers\end{tabular} &
  \begin{tabular}[c]{@{}l@{}}CNN\\ 71 (483) layers\end{tabular} \\ \hline
\multirow{3}{*}{H2O} &
  Clean &
  \begin{tabular}[c]{@{}l@{}}StackedEnsemble\\ BestOfFamily\\ 6 models\\ (AllModels\\ 75 models)\end{tabular} &
  \begin{tabular}[c]{@{}l@{}}StackedEnsemble\\ BestOfFamily\\ 6 models\\ (AllModels\\ 192 models)\end{tabular} &
  \begin{tabular}[c]{@{}l@{}}StackedEnsemble\\ AllModels\\ 296 (341)\\ models\end{tabular} &
  \begin{tabular}[c]{@{}l@{}}XGBoost\\ (StackedEnsemble\\ AllModels\\ 6 models)\end{tabular} &
  \begin{tabular}[c]{@{}l@{}}XGBoost\\ (StackedEnsemble\\ AllModels\\ 15 models)\end{tabular} &
  \begin{tabular}[c]{@{}l@{}}StackedEnsemble\\ AllModels\\ 25 (30)\\ models\end{tabular} \\ \cline{2-8} 
 &
  Noise &
  \begin{tabular}[c]{@{}l@{}}StackedEnsemble\\ AllModels\\ 72 models\end{tabular} &
  \begin{tabular}[c]{@{}l@{}}StackedEnsemble\\ AllModels\\ 91 models\end{tabular} &
  \begin{tabular}[c]{@{}l@{}}StackedEnsemble\\ AllModels\\ 135 models\end{tabular} &
  \begin{tabular}[c]{@{}l@{}}StackedEnsemble\\ SE AllModels\\ 3 models\end{tabular} &
  \begin{tabular}[c]{@{}l@{}}StackedEnsemble\\ BestOfFamily\\ 4 models\end{tabular} &
  \begin{tabular}[c]{@{}l@{}}StackedEnsemble\\ AllModels\\ 28 models\end{tabular} \\ \cline{2-8} 
 &
  Rotation &
  \multicolumn{3}{c|}{StackedEnsemble BestOfFamily 6 models} &
  XGBoost &
  \begin{tabular}[c]{@{}l@{}}StackedEnsemble\\ AllModels\\ 4 models\end{tabular} &
  \begin{tabular}[c]{@{}l@{}}StackedEnsemble\\ AllModels\\ 24 models\end{tabular} \\ \hline
\multirow{3}{*}{TPOT} &
  Clean &
  \begin{tabular}[c]{@{}l@{}}LogisticReg.\\ +DT clf\\ +KNN clf\\ (same)\end{tabular} &
  \begin{tabular}[c]{@{}l@{}}RF clf+2 models\\ +KNN clf\\ (GB clf+2 models\\ +KNN clf)\end{tabular} &
  \begin{tabular}[c]{@{}l@{}}GB clf\\ +KNN clf\\ (same)\end{tabular} &
  \multicolumn{3}{c|}{RF clf (same)} \\ \cline{2-8} 
 &
  Noise &
  \multicolumn{2}{c|}{KNN clf} &
  \begin{tabular}[c]{@{}l@{}}MultinomialNB+\\ KNN clf\end{tabular} &
  \multicolumn{2}{c|}{LinearSVC} &
  \begin{tabular}[c]{@{}l@{}}OneHotEncoder\\ +KNN clf\end{tabular} \\ \cline{2-8} 
 &
  Rotation &
  \begin{tabular}[c]{@{}l@{}}GB clf\\ +RF clf\\ +KNN clf\end{tabular} &
  \begin{tabular}[c]{@{}l@{}}ET clf\\ +KNN clf\end{tabular} &
  \begin{tabular}[c]{@{}l@{}}GB clf\\ +5 models\\ +KNN clf\end{tabular} &
  \multicolumn{2}{c|}{KNN clf} &
  RF clf \\ \hline
\end{tabular}%
}
\end{table}

Let's then move our focus to the columns with clean training data and dirty testing data in Table \ref{tab:accuracies_summary}. With the Digits dataset, TPOT seems to beat H2O when the test data fault source is noise and also when it's rotation. Thus it is interesting to see that the accuracies are pretty even with the Fashion dataset although in Figure \ref{fig:accuracy_plot_noise_clean_dirty} there seems to be a wide difference. We have to remember that the values in Table \ref{tab:accuracies_summary} are the best values among the three benchmarks with noise as the data fault source for each model. The difference can be explained by looking at Table \ref{tab:pipelines_summary}. As we can see, in shorter Fashion benchmarks with noise as the data fault source, H2O sometimes has only time to run the XGBoost models, which seems to give better accuracies at higher test data fault levels with noise as the data fault source \cite{repo}, thus being more robust than the longer Stacked Ensemble pipelines. With the small Digits dataset, H2O seemed to have ample time.

Moving again to the Fashion dataset, in Table \ref{tab:accuracies_summary} AutoKeras' different versions seem to perform worse when compared to H2O and TPOT when the data fault source is noise. Especially AutoKeras GPU with image augmentation enabled shows poor performance. We can see in Figure \ref{fig:accuracy_plot_noise_clean_dirty} that this is true even at the higher test data fault levels. In the same Table, all of the benchmarked systems seem to perform equally with rotation as the data fault source. Though in Figure \ref{fig:accuracy_plot_rotation_clean_dirty} we can see that AutoKeras GPU with image augmentation seems to pull ahead of the competition at higher test data fault levels. This is probably due to that the image augmentation process includes some rotations, as mentioned in Section \ref{sec:autokeras}.

\subsubsection{Pipelines}\label{sec:results_clean_pipelines}

The results for the optimal pipelines can be found in Table \ref{tab:pipelines_summary}. When inspecting only the rows with clean training data we can see a few things. Looking at the pipelines for AutoKeras' different versions and the model summaries in the repo \cite{repo}, we can see that it prefers very similar pipelines in different benchmarks, which are shared even between different versions of AutoKeras used in our tests. It also seems to mainly use one of three base pipelines of different lengths. As the training time increases, the possible modifications to these pipelines and the hyperparameters seem to appear at the end, and with the longer benchmarks, we can see a few layers being added to the end of the base pipelines. Looking at H2O's pipelines we can see that the length of the pipeline varies even more than with AutoKeras, ranging from 6 to 341 models for the Digits dataset and from 1 to 30 models for the Fashion dataset.

For H2O, we can see some of the Stacked Ensemble models explained in Section \ref{sec:h2o} and some pipelines based on a single model. The presence of a single XGBoost classifier among the StackedEnsemble models can be explained by looking at the H2O log files in our repo \cite{repo}, which show that H2O moves to the other base models discussed in Section \ref{sec:h2o} only after all base XGBoost models have been tested. Also, the training of all the XGBoost models takes most of the runtime. So if the benchmark time is limited, the XGBoost classifier could be the only option. 
For TPOT the choice of the dataset seems to affect the chosen pipeline. With the Digits dataset, TPOT seems to like the K-Neighbors classifier. With the Fashion dataset, Random Forest classifier seems to be the only choice.

\subsection{How good are AutoML generated models with dirty training data?}\label{sec:results_dirty}

\subsubsection{Accuracies}\label{sec:results_dirty_accuracies}

Let's first consider the columns with dirty training data and clean testing data in Table \ref{tab:accuracies_summary}. With the Digits dataset, H2O and TPOT seem to perform quite similarly with both data fault sources when training with dirty data, using parameters explained in the beginning of Section \ref{sec:results}. With the Fashion dataset and rotation as the data fault source, AutoKeras GPU with image augmentation is the clear winner as can be seen in Table \ref{tab:accuracies_summary}, but unfortunately seems to be the worst with noise.

\begin{figure}[!ht]
    \centering
    \begin{subfigure}[b]{0.45\textwidth}
        \centering
        \includegraphics[width=\textwidth]{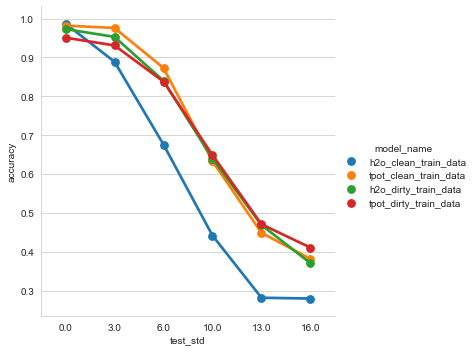}
        \caption{With noise as the data fault source.}
        \label{fig:accuracy_plot_noise_digits}
    \end{subfigure}
    \begin{subfigure}[b]{0.45\textwidth}
        \centering
        \includegraphics[width=\textwidth]{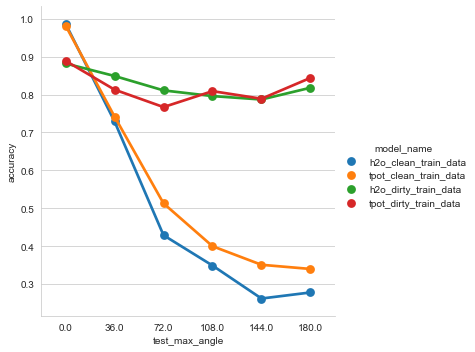}
        \caption{With rotation as the data fault source.}
        \label{fig:accuracy_plot_rotation_digits}
    \end{subfigure}
    \hfill
    \caption{Accuracy plots for the 1 h benchmark with the Digits dataset when testing at different data fault levels given the data fault source.}
\end{figure}

Let's then compare the five AutoML systems when both training and testing is done with dirty data. With the Digits dataset, H2O and TPOT seem to perform quite similarly with both data fault sources when testing with dirty data at different data fault levels, as can be seen in Figures \ref{fig:accuracy_plot_noise_digits} and \ref{fig:accuracy_plot_rotation_digits} with maybe H2O having a slight edge at mid-levels when rotation is the source.

When looking at the same results for the Fashion dataset and rotation in Figure \ref{fig:accuracy_plot_rotation_dirty_dirty}, we can see that the accuracies do not really drop as the data fault level increases as all possible rotations are covered with the huge set of training data. AutoKeras GPU with image augmentation seems to be the clear winner here while TPOT clearly performs the worst. With noise, all AutoKeras versions seem to struggle at higher data fault levels, where TPOT seems to excel, as can be seen in Figure \ref{fig:accuracy_plot_noise_dirty_dirty}. Furthermore, AutoKeras GPU with image augmentation has a peculiar performance. The accuracy on the test data seems to peak at the level that was used on the training data, clearly lacking robustness with bad scores at both ends.

\begin{table}[]
\centering
\caption{Summary of the effect of time to accuracy when both training and testing with dirty data.}
\label{tab:accuracy_time_dirty_summary}
\resizebox{0.7\textwidth}{!}{%
\begin{tabular}{|l|l|l|l|l|l|l|l|}
\hline
\multirow{2}{*}{} &
  \multicolumn{1}{c|}{\multirow{2}{*}{\begin{tabular}[c]{@{}c@{}}Training data\\ fault source\end{tabular}}} &
  \multicolumn{3}{c|}{Digits} &
  \multicolumn{3}{c|}{Fashion} \\ \cline{3-8} 
 &
  \multicolumn{1}{c|}{} &
  \multicolumn{1}{c|}{15 min} &
  \multicolumn{1}{c|}{30 min} &
  \multicolumn{1}{c|}{1 h} &
  1.5 h &
  3 h &
  6 h \\ \hline
\multirow{2}{*}{AutoKeras CPU}           & Noise    & -     & -     & -     & 0.782 & 0.819 & 0.815 \\ \cline{2-8} 
                                         & Rotation & -     & -     & -     & 0.725 & 0.795 & 0.807 \\ \hline
\multirow{2}{*}{AutoKeras GPU}           & Noise    & -     & -     & -     & 0.807 & 0.812 & 0.812 \\ \cline{2-8} 
                                         & Rotation & -     & -     & -     & 0.810 & 0.821 & 0.829 \\ \hline
\multirow{2}{*}{AutoKeras GPU with Aug.} & Noise    & -     & -     & -     & 0.837 & 0.851 & 0.845 \\ \cline{2-8} 
                                         & Rotation & -     & -     & -     & 0.860 & 0.848 & 0.877 \\ \hline
\multirow{2}{*}{H2O}                     & Noise    & 0.842 & 0.840 & 0.840 & 0.796 & 0.799 & 0.821 \\ \cline{2-8} 
                                         & Rotation & 0.820 & 0.838 & 0.818 & 0.781 & 0.788 & 0.792 \\ \hline
\multirow{2}{*}{TPOT}                    & Noise    & 0.827 & 0.827 & 0.838 & 0.776 & 0.776 & 0.782 \\ \cline{2-8} 
                                         & Rotation & 0.818 & 0.853 & 0.844 & 0.737 & 0.737 & 0.760 \\ \hline
\end{tabular}%
}
\end{table}

The results for the effect of benchmarking time to the accuracy when both training and testing with dirty data can be seen in Table \ref{tab:accuracy_time_dirty_summary}. With the Digits dataset, while H2O's scores seem to have stabilized after 15 min, TPOT might need more time to reach the optimal results. With the Fashion dataset, 1.5 h clearly is not enough for AutoKeras CPU. This can also be seen from Table \ref{tab:pipelines_summary}, which shows that after 1.5 h, the CNN has only 16 layers. The rest of the tested systems show minor improvements with time.

\subsubsection{Pipelines}\label{sec:results_dirty_pipelines}

\begin{table}[]
\centering
\caption{Summary of the optimal pipelines for TPOT given the dataset and training data fault source and level. }
\label{tab:tpot_pipelines}
\resizebox{0.9\textwidth}{!}{%
\begin{tabular}{|l|l|l|l|l|l|l|l|}
\hline
\multicolumn{1}{|c|}{\multirow{2}{*}{Dataset}} &
  \multicolumn{1}{c|}{\multirow{2}{*}{\begin{tabular}[c]{@{}c@{}}Training data\\ fault source\end{tabular}}} &
  \multicolumn{6}{c|}{Data fault level} \\ \cline{3-8} 
\multicolumn{1}{|c|}{} &
  \multicolumn{1}{c|}{} &
  0 (clean) &
  1 &
  2 &
  3 &
  4 &
  5 \\ \hline
\multirow{2}{*}{Digits} &
  Noise &
  \begin{tabular}[c]{@{}l@{}}GB clf\\ +KNN clf\end{tabular} &
  KNN clf &
  \begin{tabular}[c]{@{}l@{}}MultinomialNB\\ +KNN clf\end{tabular} &
  MultinomialNB &
  \begin{tabular}[c]{@{}l@{}}MultinomialNB\\ +KNN clf\end{tabular} &
  \begin{tabular}[c]{@{}l@{}}ET clf\\ +MultinomialNB\end{tabular} \\ \cline{2-8} 
 &
  Rotation &
  \begin{tabular}[c]{@{}l@{}}GB clf\\ +KNN clf\end{tabular} &
  \begin{tabular}[c]{@{}l@{}}GB clf\\ +MultinomialNB\\ +KNN clf\end{tabular} &
  \begin{tabular}[c]{@{}l@{}}GB clf\\ +KNN clf\end{tabular} &
  \begin{tabular}[c]{@{}l@{}}GB clf\\ +KNN clf\end{tabular} &
  \begin{tabular}[c]{@{}l@{}}RF clf\\ +ET clf\\ +KNN clf\end{tabular} &
  \begin{tabular}[c]{@{}l@{}}GB clf\\ +5 models\\ +KNN clf\end{tabular} \\ \hline
\multirow{2}{*}{Fashion} &
  Noise &
  RF clf &
  XGB clf &
  \begin{tabular}[c]{@{}l@{}}OneHotEncoder\\ +KNN clf\end{tabular} &
  LinearSVC &
  \begin{tabular}[c]{@{}l@{}}LinearSVC\\ + GB clf\end{tabular} &
  LinearSVC \\ \cline{2-8} 
 &
  Rotation &
  \multicolumn{6}{c|}{RF clf} \\ \hline
\end{tabular}%
}
\end{table}

When comparing the pipelines that the tested systems produce, we can see from Table \ref{tab:pipelines_summary}, and from the model summaries in the repo \cite{repo}, that the general pipelines for AutoKeras' versions and H2O do not change that much, even though H2O has its issues with the large dataset and short benchmark time. With TPOT, the preferred pipelines tend to change a lot more based on the dataset and the data fault source. With the smaller Digits dataset, if rotation is the source, TPOT seems to favor the K-Neighbors classifier as part of the pipeline as can be seen in Table \ref{tab:tpot_pipelines}. This is also true with noise as the source if the data fault level is low. With higher levels of noise in the training data, Multinomial Naive Bayes seems to be the preferred choice. Regarding the larger Fashion dataset, with noise as the data fault source, Logistic Regression seems to be the model of choice at higher data fault levels. On the other hand, with rotation, TPOT seems to use a Random Forest classifier at all levels.

\subsection{Recommendations and comparison}\label{sec:results_recommendations}

To begin with, the following recommendations for different use cases can be made, based on the results above:

\begin{itemize}

\item[--]

When both training and testing with clean data, AutoKeras GPU with image augmentation seems to be the clear winner but requires a lot of time and computing power.

\item[--]

When training with clean data and testing with dirty data, due to the inconsistencies of H2O with shorter training times, TPOT would be the optimal choice when noise is the data fault source. With rotation, AutoKeras GPU with image augmentation would be the top choice because of good performance with both clean and very faulty test data.  

\item[--]

When training with dirty data and testing with clean data, while AutoKeras GPU with image augmentation is the clear winner with rotation, whereas with noise there is no clear winner. H2O seems to best TPOT with both datasets, but equal the performance of the two other AutoKeras versions with Fashion.

\item[--]

When both training and testing with dirty data, TPOT seems to be the winner when noise is the data fault source because of its constantly good performance even at high data fault levels. AutoKeras GPU with image augmentation is once again the clear winner with rotation.

\end{itemize}

Given the test data fault source, with the larger datasets like Fashion, where good training images are plenty, training with dirty data is in most cases the better option with both data fault sources as can be seen when comparing the plots for each model in Figures \ref{fig:accuracy_plot_noise_clean_dirty} and \ref{fig:accuracy_plot_noise_dirty_dirty} for noise, and Figures \ref{fig:accuracy_plot_rotation_clean_dirty} and \ref{fig:accuracy_plot_rotation_dirty_dirty} for rotation. This is also the case with the much smaller Digits dataset when using rotation as the data fault source, as can be seen in Figure \ref{fig:accuracy_plot_rotation_digits}. However, with noise, training with dirty data is not necessarily the best option as can be seen in Figure \ref{fig:accuracy_plot_noise_digits}. In fact, training with dirty data seems to be the clear winner only in H2O's case. With TPOT the models seem to perform quite similarly at the mid and higher data fault levels. The other exception to this rule, is obviously when we know that the test data is clean.

\subsection{Resource usage}\label{sec:usage}

There were several differences in the resource usage between the three AutoML systems. With 40 cores AutoKeras used almost $100\%$ of the CPU resources available and around 8 GB of RAM. The two GPU versions of AutoKeras used almost exclusively GPU and the same amount of memory.

H2O's CPU usage was around $80\%$ It typically used around 150GB with the larger Fashion dataset when given 350GB of memory to use for the Java process. Some of the runs failed due to a segmentation fault and had to be rerun.

We noticed that TPOT had trouble parallelizing some of the models it uses, and CPU efficiency was around $30\%$. We observed times when only one core's usage was maxed out. Regarding the RAM usage, for TPOT around 30 GB was usually enough.

\section{Discussion}\label{sec:disc}

The authors of Fashion-MNIST report accuracy scores, with a similar train-test split, for several classifiers and state-of-the-art Neural Networks (NNs). \footnote{\url{https://github.com/zalandoresearch/fashion-mnist}} The best reported score was 0.897 for basic classifiers (SVC) and 0.967 for state-of-the-art NNs (WRN40-4). Looking at our best results with clean test data in Table \ref{tab:accuracies_summary}, even with dirty training data our best results compare quite well.

Regarding TPOT's memory usage mentioned in Section \ref{sec:usage}, the official documentation of TPOT includes a warning of possible memory issues when multiple cores are used \cite{OlsonGECCO2016}. We too noticed occasional peaks to around 200 GB when testing with 96 cores. Others \footnote{\url{https://github.com/szilard/GBM-multicore}} have also noticed issues with ML systems when using too many cores.

To explain the results from Section \ref{sec:results_recommendations} related to the similar performance of models trained with clean or dirty data with the smaller Digits dataset and noise, we have to consider the nature of the data fault sources. When using data fault source like rotation, most information in the image is retained in the data even at higher data fault levels. However, a data fault source like Gaussian noise destroys parts of the information in the image, as can be seen in Figure \ref{fig:digits_data_fault_noise}. Because the Digits dataset is small and the random nature of Gaussian noise, we are left with only a few good training images, so training with dirty data may not be the best choice after all. Also, the effects of Gaussian noise are particularly noticeable with the Digits dataset, because we are using very low-resolution images as training data. Regarding the larger Fashion dataset, there are still plenty good training images within the huge training dataset.

As for the results in \ref{sec:results_clean} regarding AutoKeras' bad performance with clean training data and dirty test data, this is a known problem for neural networks as discussed in Section \ref{sec:intro}. In the image augmentation enabled version's case, it could be said that because of the random crops, horizontal flips, and cutouts discussed in Section \ref{sec:autokeras}, the neural network becomes even more sensitive to certain data fault sources destroying information from the data.

It is also known that AutoML systems can recommend alternative optimal solutions in different runs for the same problem \footnote{\url{https://epistasislab.github.io/tpot/using/}}. We also observed this. The cause may be the tight time limits imposed on a system with stochastic elements or just that two pipelines offer almost equal performance.

\section{Conclusions}\label{sec:conclusions}

Based on the results, using training data, which contains examples of faults the system will encounter is promising: accuracy with clean test data drops a bit but robustness increases a lot. We also noted that different AutoML systems produce very different ML pipelines. TPOT even generated rather different pipelines for clean, noisy, and rotated data. 

Future work of exploring if our findings apply to not so similar datasets and different data fault sources is important. Future AutoML tools may want consider robustness as an explicit optimization goal. Perhaps the user could specify preferred trade-off between accuracy and robustness.

\bibliographystyle{eptcs}
\bibliography{refs}

\end{document}